\documentclass{article}
\usepackage{spconf,amsmath,amssymb,graphicx}
\usepackage[hyphens]{url}
\usepackage{booktabs}
\usepackage{caption}
\usepackage{placeins}

\makeatletter
\renewcommand{\paragraph}[1]{\par\medskip\noindent\textbf{#1}\hspace{0.5em}\ignorespaces}
\makeatother

\title{AnchorMTS: Multi-View Signal Representation and Forecast-Anchored Fusion for Long-Term Time-Series Forecasting}
\name{Gao Wu$^{1}$\enspace Tingting Xu$^{2,3}$\enspace Changning Wu$^{1}$\enspace Tengfei Long$^{1}$\enspace Yong Liu$^{1}$\enspace Jiabin Jin$^{2,3,*}$\enspace Kexin Zhang$^{1,*}$
  \thanks{$^{*}$Corresponding authors: Jiabin Jin (jinjiabin1987@126.com), Kexin Zhang (zhangkexin@zju.edu.cn).}}
\address{%
  \normalsize $^{1}$College of Control Science and Engineering, Zhejiang University, Hangzhou, China \\
  \normalsize $^{2}$Zhejiang Provincial Security Science and Technology Research Institute, Hangzhou, China \\
  \normalsize $^{3}$Zhejiang Province Key Laboratory of ``Public Security Brain'' Big Data and Key Security Technologies Research, Hangzhou, China}

\begin{document}
\ninept
\newcommand{\FinalMacroMSE}{0.336}
\newcommand{\FinalMacroMAE}{0.367}
\newcommand{\FinalEnvelopeMSECI}{[-0.0071, -0.0016]}
\newcommand{\FinalEnvelopeMSEP}{0.0227}
\newcommand{\FinalTemporalMSE}{0.340}
\newcommand{\FinalTemporalMAE}{0.369}
\newcommand{\FinalTemporalDeltaMSE}{-0.0043}
\newcommand{\FinalTemporalDeltaMAE}{-0.0019}
\newcommand{\FinalTemporalMSEWL}{19/5}
\newcommand{\FinalTemporalMSECI}{[-0.0068, -0.0021]}
\newcommand{\FinalTemporalMSEP}{0.0066}
\newcommand{\FinalAlignedMSE}{0.3164}
\newcommand{\FinalShuffledMSE}{0.3171}
\newcommand{\FinalShuffleDeltaMSE}{+0.00078}
\newcommand{\FinalShuffleMSEWL}{12/6}
\newcommand{\FinalShuffleMSECI}{[-0.00013, +0.00164]}
\newcommand{\FinalShuffleMSEP}{0.2379}
\newcommand{\FinalActivationPercent}{4.2115\%}
\newcommand{\FinalForecastCount}{1,663,840}
\newcommand{\FinalOriginalRatioP}{0.0451/0.2299/8.1402}
\newcommand{\FinalBoundedRatioP}{0.0451/0.2299/0.2500}
\newcommand{\FinalBoundMSEChange}{-4.7\times 10^{-5}}
\newcommand{\FinalImageShuffleDelta}{0.002169}
\newcommand{\FinalJointShuffleDelta}{0.002162}
\newcommand{\DirectImageShuffleDelta}{0.005309}
\newcommand{\DirectJointShuffleDelta}{0.005302}

\maketitle
\begin{abstract}
Long-term time-series forecasters typically optimize a single numerical representation, even though the same history exposes complementary geometry, recurrence, and semantic structure. We propose AnchorMTS, which derives self-aligned visual and textual views from each observed window and uses them only to refine a complete numerical forecast. A trainable evidence path produces a candidate forecast, while a parameter-free output projection enforces an exact per-variable bound on its displacement from the numerical anchor. Across six benchmarks and four horizons, AnchorMTS achieves the best macro MSE/MAE (0.336/0.367) among seven supervised baselines and wins 18 of 24 MSE settings against the cellwise baseline envelope; removing the auxiliary path raises macro MSE to 0.340. The projection leaves 95.8\% of candidate refinements unchanged while yielding zero bound violations, showing that complementary multi-view evidence can improve forecasting while its final influence remains explicitly controlled.

\end{abstract}
\section{Introduction}
\label{sec:intro}

Long-term time-series forecasting (LTSF) supports planning in energy, weather, transportation, and finance. Modern supervised methods reorganize numerical histories through decomposition \cite{wu2021autoformer,zhou2022fedformer}, patching \cite{nie2022patchtst}, two-dimensional temporal variation \cite{wu2022timesnet}, inverted variable tokens \cite{liu2024itransformer}, multiscale mixing \cite{wang2024timemixer}, or endogenous--exogenous interaction \cite{wang2024timexer}. These advances are substantial, yet most numerical LTSF models ultimately optimize forecasting within a single learned representation space.

The same observed history admits complementary representations without adding measurements. Gramian angular fields (GAFs) expose pairwise angular relations \cite{wang2015imaging}; line renderings make slopes and turning points spatially local; recurrence plots reveal repeated states \cite{eckmann1995recurrence}; and a compact description summarizes trend, periodicity, and variable-level stability. Frozen vision--language encoders can organize such views, but they were not trained as forecasting models. This creates a central tension: alternative representations may reveal useful structure, yet auxiliary representations should not be allowed to replace a strong numerical prediction.

Recent multimodal forecasters demonstrate that alternative views can complement numerical sequences. VisionTS formulates forecasting as visual reconstruction, Time-VLM couples temporal, visual, and textual learners, DMMV integrates numerical and visual predictors, and TimesCLIP aligns visual and textual representations \cite{chen2024visionts,siru2025time,shen2026multi,dong2025teaching}. Related multi-view formulations also explore adaptive tri-modal fusion, time/frequency/vision representations, evidence-theoretic aggregation, and uncertainty-aware view weighting \cite{chowdhury2026t3time,ao2026trits,zhan2025time,kamarthi2022camul}. AnchorMTS addresses a complementary question in full-data supervised LTSF: how can history-derived evidence refine a strong numerical forecast without taking control of the prediction? Rather than treating derived views as peer forecasters or replacements for the numerical representation, AnchorMTS uses them only to propose residual evidence around an independently complete numerical forecast. Our formulation separates two questions: whether auxiliary representations can improve the numerical prediction, and how much authority they should have over the final forecast. The trainable evidence path addresses the first, while the exact forecast-space projection addresses the second.

AnchorMTS realizes this principle with two coupled forecasting paths. A numerical backbone produces a complete anchor forecast, while frozen image and text representations condition a trainable evidence path that generates a candidate refinement. Intermediate updates are norm-controlled, and a final parameter-free projection constrains the candidate directly around the numerical anchor in forecast space.

Our contributions are threefold. First, we introduce a self-derived multi-view signal interface that exposes angular, trajectory, recurrence, and semantic structure without external measurements or manual alignment. Second, we formulate multimodal forecasting as bounded refinement of a complete numerical predictor, separating learned evidence utility from explicit final forecast authority. Third, we validate AnchorMTS across six benchmarks and four horizons, demonstrating competitive forecasting gains, complementary frozen-view structure, and an exact forecast-space influence bound.

\section{Method}
\label{sec:method}

\begingroup
\renewcommand{\baselinestretch}{0.91}\small

Given an observed window $X\in\mathbb{R}^{L\times C}$, the task is to predict $Y\in\mathbb{R}^{H\times C}$. AnchorMTS builds two paths from $X$: a primary numerical forecaster and an auxiliary representation path. Figure~\ref{fig:method} makes their roles explicit.

\begin{figure*}[t]
\centering
\includegraphics[width=\textwidth]{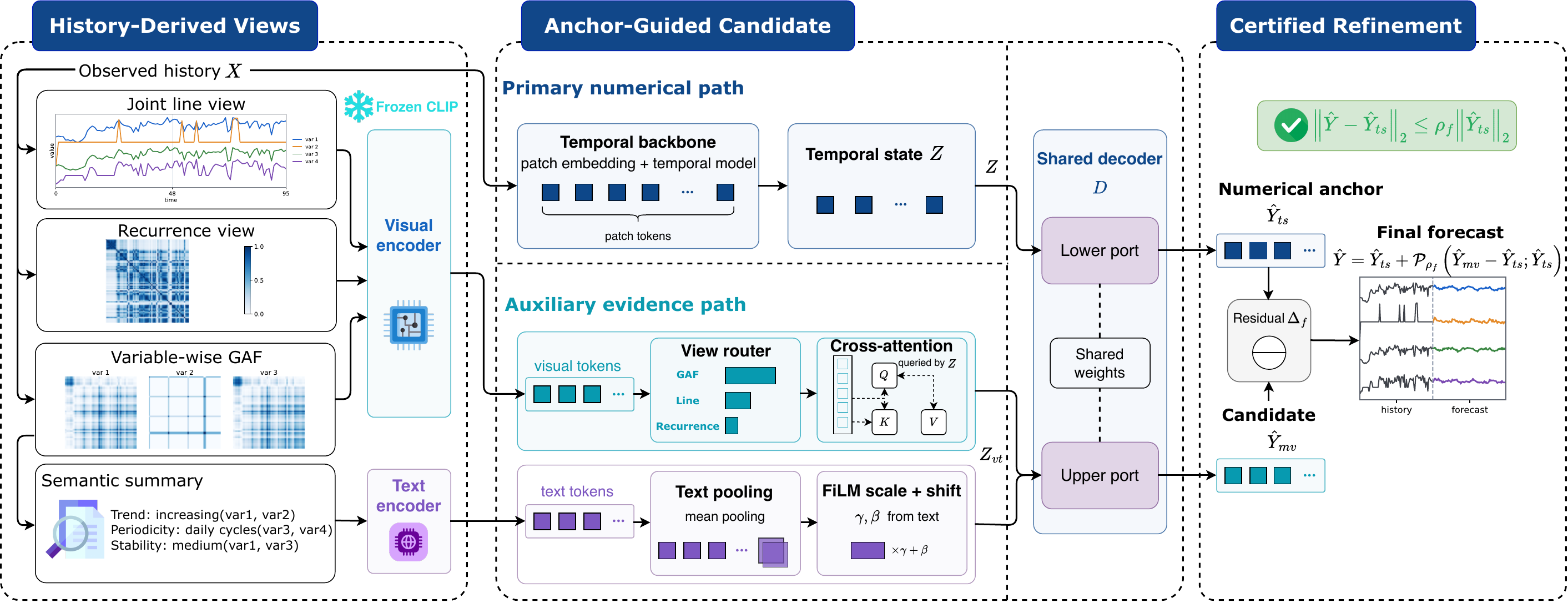}
\caption{Overview of AnchorMTS. The observed history is processed by a complete numerical forecaster to produce the anchor prediction, while self-derived angular, trajectory, recurrence, and textual views provide cached auxiliary evidence. Evidence fusion proposes a multimodal candidate, and an exact forecast-space projection converts its residual into a bounded refinement of the numerical anchor.}
\label{fig:method}
\end{figure*}

\subsection{Self-Derived Representations}

For variable $c$, min--max scaling maps the history to $\tilde x_i^c\in[-1,1]$ and $\phi_i^c=\arccos(\tilde x_i^c)$. The GAF $G_{ij}^c=\cos(\phi_i^c+\phi_j^c)$ is a deterministic pairwise angular re-encoding; channel-labelled tiles preserve variable identity. With $z_i^c$ denoting within-window standardized values and $s_i=C^{-1}\sum_c z_i^c$, the line view renders the ordered path $(i,s_i)$, while
\[
R_{ij}=\exp\!\left(-|s_i-s_j|/(\sigma+\epsilon)\right)
\]
forms a recurrence view, where $\sigma$ is the median positive pairwise distance. A fixed prompt asks GPT-4.1-mini to summarize only the historical GAF's trend, periodicity, stability, and variable dynamics. Thus the three images and text contain no future values or external measurements.

Frozen CLIP ViT-B/32 towers encode each view once \cite{radford2021learning}. Their last-layer sequences---50 tokens per image and at most 77 text tokens---are cached and reused across training epochs and horizons. These views intentionally expose different locality structures of the same observed history. The variable-wise GAF preserves channel identity while turning pairwise angular relations into spatial patterns; the line view makes ordered slopes and turning points locally accessible; the recurrence view emphasizes returns to similar states irrespective of their absolute timestamps; and the caption compresses variable-level trend, periodicity, and stability into a semantic summary. Their purpose is therefore not to introduce additional measurements, but to make complementary structures of $X$ accessible to frozen representation encoders before fusion.

We deliberately retain channel separation in the GAF to preserve cross-variable heterogeneity, whereas the line and recurrence views operate on the standardized joint trajectory to summarize shared temporal evolution without multiplying the number of visual streams by $C$.

The representation interface is deliberately fixed with respect to forecasting optimization. The visual and textual encoders are frozen, and all four views are aligned by the same observed-window index before training the forecasting path. Consequently, the forecasting model learns how to use a stable auxiliary representation space rather than jointly reshaping the pretrained encoders for each dataset or horizon. The same cached evidence can therefore be reused across epochs and prediction horizons while the trainable numerical and fusion paths adapt to the forecasting objective.

A lightweight router assigns sample-dependent weights to the three visual memories before evidence fusion.

\subsection{Forecast-Anchored Evidence Fusion}

The numerical backbone maps $X$ to temporal tokens $Z\in\mathbb{R}^{N\times d}$, including patch and variable summaries, and a decoder $D$ produces the complete anchor forecast $\widehat Y_{\rm ts}$. Let $E_v^{(k)}$ be the cached CLIP sequence for visual view $k\in\{\mathrm{GAF,line,RP}\}$. Projection and a learned view identity give
\begin{equation}
V^{(k)}=\operatorname{Proj}_v(E_v^{(k)})+e^{(k)},\qquad
\alpha=\operatorname{softmax}_k r\!\left(\operatorname{mean}V^{(k)}\right).
\end{equation}
A shared layer-normalized linear scorer maps the pooled representation of each visual view to a softmax weight, producing the routed visual memory $M_v$.
For the $K$ available views, the routed visual memory is
$M_v=\operatorname{concat}_k(K\alpha_kV^{(k)})$. The routed memory therefore preserves the full token sequence of each available view while adapting its contribution to the current history.

The temporal tokens query this memory through cross-attention,
\begin{equation}
\Delta_v=\operatorname{MHA}(\operatorname{LN}Z,\operatorname{LN}M_v,\operatorname{LN}M_v).
\end{equation}
For any proposed update $\Delta$ around state $A$, define the token-wise norm control
\[
\mathcal{B}_{\rho}(\Delta;A)=\Delta\min\!\left(1,
\frac{\rho\lVert A\rVert_2}{\lVert\Delta\rVert_2+\epsilon}\right).
\]
With a learned element-wise gate $g_v$, the visual state is
$Z_v=Z+g_v\odot\mathcal{B}_{\rho_v}(\Delta_v;Z)$, where $\rho_v=0.60$.

Text is conditioned separately. If $E_t$ and $m$ are the cached text tokens and attention mask, masked pooling gives
\begin{equation}
t=\frac{\sum_jm_j\operatorname{Proj}_t(E_{t,j})}{\sum_jm_j},\qquad
(s,b)=\operatorname{FiLM}(t).
\end{equation}
The proposal $\Delta_t=Z_v\odot s+b$ is bounded and gated as
$Z_{vt}=Z_v+g_t\odot\mathcal{B}_{\rho_t}(\Delta_t;Z_v)$ with $\rho_t=0.25$.

Visual and textual evidence play deliberately different roles in the fusion path. The three visual representations retain token-level spatial structure, so the temporal state queries their routed memory through cross-attention. The caption instead provides a compact global description of trend, periodicity, and variable behavior; it therefore bypasses the visual router and modulates the visually conditioned state through feature-wise scale and shift. This asymmetric design preserves fine-grained visual evidence for token-level retrieval while using text as a global semantic conditioning signal.

The intermediate budgets $\rho_v=0.60$ and $\rho_t=0.25$ are fixed representation-space regularization settings shared across all datasets and horizons; they are not the source of the forecast-level guarantee.
The shared decoder maps the untouched temporal state to the anchor forecast $\hat{Y}_{ts}$, while the evidence-conditioned state enters the learned forecast-residual branch to produce the multimodal candidate.

Using the same decoder for both states is important to the anchor formulation: the numerical anchor and multimodal candidate share the same forecast parameterization and differ only through the evidence-conditioned state and residual branch. The resulting forecast difference can therefore be interpreted directly as an evidence-induced correction before the final output constraint is applied.

The anchor and candidate share the same inverse normalization and channel-wise affine output map, yielding two forecasts in the reported prediction space before the final projection. The visual and textual state controls shape this candidate in representation space; the exact forecast-level constraint is imposed separately on the final outputs.

\subsection{Exact Final-Output Projection}

Define the candidate correction $\Delta_f=\widetilde Y_{\rm mv}-\widehat Y_{\rm ts}$. The final reported forecast is
\begin{align}
\widehat Y&=\widehat Y_{\rm ts}+\mathcal{P}_{\rho_f}(\Delta_f;\widehat Y_{\rm ts}),\\
\mathcal{P}_{\rho}(\Delta;A)&=\Delta\min\!\left(1,
\frac{\rho\lVert A\rVert_2}{\lVert\Delta\rVert_2+10^{-8}}\right),
\end{align}
We fix $\rho_f=0.25$ globally, defining a 25\% relative horizon-wise $L_2$ budget around the numerical anchor for every sample--variable forecast. The norm is computed across the $H$ forecast steps, giving
\[
\lVert\widehat Y-\widehat Y_{\rm ts}\rVert_2
\leq \rho_f\lVert\widehat Y_{\rm ts}\rVert_2
\]
by construction.

The operator is the identity whenever the candidate correction already lies inside the prescribed ball. Otherwise, it preserves the correction direction and rescales only its magnitude to the boundary. Thus the constraint does not define a new forecast direction; it only limits the distance that the evidence-conditioned candidate may move from the complete numerical anchor.

Because $\mathcal{P}_{\rho_f}$ is applied to the two fully formed forecasts after their shared output map, the bound is independent of the internal representation geometry and is directly expressed in the units used for evaluation.
\endgroup

\section{Experiments}
\label{sec:experiments}

\begingroup
\renewcommand{\baselinestretch}{0.91}\small

\begin{figure}[!b]
\centering
\includegraphics[width=0.99\columnwidth]{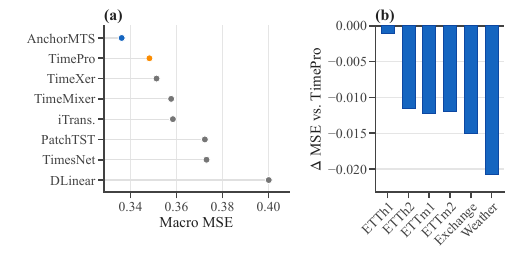}
\caption{Forecasting accuracy. (a) Macro MSE over the 24 aligned dataset-horizon cells. (b) Dataset-average MSE difference from TimePro, the strongest individual baseline by macro error; negative values favor AnchorMTS.}
\label{fig:main_results}
\end{figure}

\subsection{Setup}

We evaluate ETTh1, ETTh2, ETTm1, ETTm2, Exchange, and Weather with input length 96 and horizons $H\!\in\{96,192,\allowbreak336,720\}$, yielding 24 cells. Standard chronological splits and training-partition normalization are used \cite{zhou2021informer,lai2018modeling,wu2021autoformer}; MSE and MAE average all forecast steps and variables. Every cell uses the same self-derived views, frozen CLIP path, fusion topology, and $(\rho_v,\rho_t,\rho_f)=(0.60,0.25,0.25)$. Ordinary backbone and optimization settings are selected per cell. We compare with seven supervised LTSF baselines spanning linear, patch, two-dimensional, inverted-token, multiscale, and endogenous--exogenous designs \cite{zeng2023dlinear,nie2022patchtst,wu2022timesnet,liu2024itransformer,wang2024timemixer,wang2024timexer,ma2025timepro}. Table~1 includes methods with complete results under this aligned 24-setting supervised protocol.

\subsection{Main Forecasting Results}

\begin{table}[t]
\centering
\scriptsize
\setlength{\tabcolsep}{2.2pt}
\renewcommand{\arraystretch}{1.45}
\begin{tabular}{lcccc}
\toprule
Method & MSE $\downarrow$ & MAE $\downarrow$ & MSE W/T/L & MAE W/T/L \\
\midrule
AnchorMTS & \textbf{0.336} & \textbf{0.367} & -- & -- \\
TimePro & 0.348 & 0.374 & 22/0/2 & 20/1/3 \\
TimeXer & 0.351 & 0.374 & 20/1/3 & 18/1/5 \\
TimeMixer & 0.358 & 0.384 & 23/0/1 & 21/1/2 \\
iTransformer & 0.358 & 0.379 & 24/0/0 & 23/1/0 \\
PatchTST & 0.372 & 0.388 & 23/0/1 & 23/0/1 \\
TimesNet & 0.373 & 0.391 & 24/0/0 & 23/0/1 \\
DLinear & 0.400 & 0.419 & 24/0/0 & 24/0/0 \\
\midrule
Cellwise best$^*$ & 0.340 & 0.369 & 18/0/6 & 15/1/8 \\
\bottomrule
\end{tabular}
\caption{Macro comparison over 24 aligned dataset--horizon settings. W/T/L compares AnchorMTS with each baseline at the reported three-decimal precision. *Cellwise best denotes the metric-wise envelope over the seven baselines rather than an additional forecasting model.}
\label{tab:main_comparison}
\end{table}

AnchorMTS achieves the lowest 24-cell macro errors, with MSE/\allowbreak MAE of 0.336/0.367 (Table~1). Against the metric-wise cellwise envelope of the seven baselines, it wins 18/24 MSE settings and 15/24 MAE settings with one MAE tie; the cell-resampled MSE-difference interval is $[-0.0071,-0.0016]$. Figure~2 shows that the dataset-average MSE gain over TimePro spans all six benchmarks. The largest margin occurs on Weather, the highest-dimensional benchmark in the grid, a pattern consistent with the stronger channel-diversity organization of the variable-wise GAF in Table~2.

The advantage also persists across forecast length. Averaging the six datasets at each horizon, AnchorMTS improves over TimePro at $H=96,192,336,$ and $720$, with MSE margins of $-0.0075$, $-0.0090$, $-0.0138$, and $-0.0181$, respectively. AnchorMTS remains favorable at all four horizons, with the largest absolute MSE margin at $H=720$. Together with the six dataset-level gains in Fig.~2(b), the improvement is distributed across both dataset identity and forecast length.

\begin{figure*}[!t]
\centering
\includegraphics[width=0.97\textwidth]{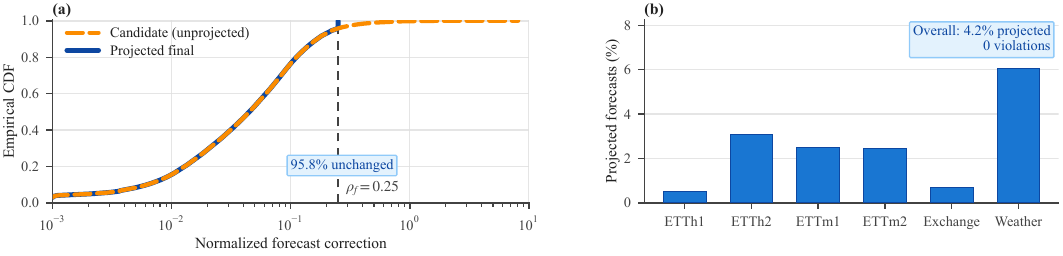}
\caption{Exact forecast-space projection. (a) Correction-ratio distributions before and after the final projection over all 24 settings; candidate refinements inside $\rho_f=0.25$ pass unchanged, while the extreme tail is truncated at the prescribed radius. (b) Fraction of sample--variable forecasts projected in each dataset. Overall, 95.8\% of refinements pass unchanged and no reported forecast violates the bound.}
\label{fig:bounded-influence}
\end{figure*}

\subsection{Auxiliary-Path Ablation}
\label{sec:controls}

We compare AnchorMTS with a configuration-matched temporal-only architectural ablation that retains the selected numerical backbone, forecasting heads, optimizer, loss, and seed but removes the auxiliary representation and fusion modules. Across all 24 settings, the complete model lowers macro MSE/MAE from 0.340/0.369 to 0.336/0.367, improves MSE in 19 settings, and yields a paired MSE-difference interval of $[-0.0068,-0.0021]$. Because the temporal backbone and its forecasting configuration are retained, this comparison evaluates the complete auxiliary pathway as an additive refinement rather than as a replacement forecaster. Improvements span datasets and horizons, not one benchmark.

\subsection{Exact Forecast-Space Bound}

Across 1,663,840 sample--variable forecasts, 95.8\% of candidate refinements already satisfy the prescribed radius and pass through unchanged. The projection preserves the median and 95th-percentile correction ratios (0.0451/0.2299), truncates the maximum from 8.1402 to 0.2500, and yields zero violations. Macro MSE changes by only $-4.7\times10^{-5}$, showing that the exact bound primarily removes rare extreme corrections without suppressing ordinary refinements.

These statistics distinguish exact influence control from generic shrinkage. Since the median and 95th-percentile corrections are unchanged, the projection does not systematically attenuate the evidence path; it intervenes only on the small tail that exceeds the prescribed forecast-space budget. Figure~3(b) further shows that activation is dataset dependent rather than uniformly imposed, while the aggregate projection rate remains only 4.2\%. Thus the candidate forecaster retains its ordinary refinements in the vast majority of cases, whereas the final reported prediction always satisfies the same interpretable constraint.

The activation profile also varies substantially across datasets: Weather exhibits the highest projection rate, whereas ETTh1 and Exchange show substantially lower activation. Because the budget is defined relative to each numerical anchor, its activation rate is free to vary across datasets rather than being fixed by a global correction threshold. Importantly, the same $\rho_f$ is used in every setting, so the dataset-dependent activation emerges from the learned candidate corrections rather than dataset-specific tuning of the final constraint.

\subsection{Complementarity of Frozen Views}

To examine whether the frozen encoders retain the structures targeted by each self-derived view, we retrieve nearest neighbors in each visual embedding space and compare numerical trend, periodicity, recurrence, and channel-diversity descriptors with arbitrary window pairs. Retrieval lift measures the relative reduction in descriptor distance.

\begin{table}[h]
\centering
\setlength{\tabcolsep}{3.0pt}
\begin{tabular}{lrrrr}
\toprule
Frozen view & Trend & Period. & Recur. & Chan. div. \\
\midrule
Var.-wise GAF & 0.107 & 0.109 & 0.109 & \textbf{0.161} \\
Joint line & \textbf{0.279} & \textbf{0.324} & \textbf{0.329} & 0.039 \\
Recurrence plot & 0.259 & 0.248 & 0.287 & 0.042 \\
\bottomrule
\end{tabular}
\caption{Retrieval lift of frozen visual representations, macro-averaged across six training sets. Higher values indicate that representation-space neighbors are more similar in the corresponding numerical signal descriptor than arbitrary window pairs.}
\label{tab:representation_retrieval}
\end{table}

Table~2 reveals complementary organization in the frozen spaces: line and recurrence views primarily preserve aggregate temporal dynamics, whereas the variable-wise GAF is strongest for cross-channel diversity. This pattern agrees with the intended roles of the self-derived representations.

The division of labor is consistent with how the views are constructed. The joint line and recurrence representations compress channels into a shared trajectory and therefore emphasize common temporal evolution, whereas the variable-wise GAF retains channel identity and exposes heterogeneous pairwise geometry. The frozen encoder therefore preserves distinct signal organization across the three view constructions, with each view emphasizing the structure it was designed to make locally accessible.

Taken together, the experiments support the three roles separated by the model design. The main benchmark and temporal ablation establish the forecasting value of the complete auxiliary path; the retrieval analysis shows that the frozen views retain complementary signal organization; and the projection analysis verifies that this additional evidence remains subject to an exact forecast-space influence budget. These results align the empirical evaluation with the representation, refinement, and control stages of Fig.~1.
\endgroup

\section{Conclusion}

AnchorMTS formulates multimodal time-series forecasting as bounded refinement of a complete numerical predictor. Self-derived visual and textual views expose complementary structure in the same observed history, while a trainable evidence path uses this auxiliary information to refine the numerical forecast. An exact output projection independently bounds the authority of the resulting correction. Across six benchmarks and four horizons, this formulation achieves the best macro error among seven supervised baselines while preserving an explicit variable-wise influence bound.

\clearpage

\appendix
\setlength{\dbltextfloatsep}{6pt plus 1pt minus 1pt}
\setlength{\dblfloatsep}{5pt plus 1pt minus 1pt}

\section{Overview}

This appendix reports the complete forecasting grid and cellwise references,
the evaluation and search protocol, exact caption-generation details, the
comparison scope of closest multimodal work, representation evidence, and
reproduction entry points.

\section{Complete Forecasting Results}

Table~\ref{tab:full_baseline_grid_main} reports every dataset--horizon cell against the seven external baselines, and Table~\ref{tab:full_grid_supp} identifies the metric-wise cellwise references. All AnchorMTS entries are exact final-output predictions.

\paragraph{Statistical tests.}
Against the metric-wise cellwise baseline envelope, the main model has exact sign-test $p=0.0227$ for MSE. Against the configuration-matched temporal-only ablation, it has exact sign-test $p=0.0066$ for MSE.

\begin{table*}[!t]
\centering
\setlength{\tabcolsep}{1.25pt}
\renewcommand{\arraystretch}{0.90}
\begin{scriptsize}
\resizebox{\textwidth}{!}{%
\begin{tabular}{llcccccccccccccccc}
\toprule
Dataset & H & \multicolumn{2}{c}{AnchorMTS} & \multicolumn{2}{c}{TimePro} & \multicolumn{2}{c}{TimeXer} & \multicolumn{2}{c}{TimeMixer} & \multicolumn{2}{c}{iTransformer} & \multicolumn{2}{c}{PatchTST} & \multicolumn{2}{c}{TimesNet} & \multicolumn{2}{c}{DLinear} \\
\cmidrule(lr){3-4}\cmidrule(lr){5-6}\cmidrule(lr){7-8}\cmidrule(lr){9-10}\cmidrule(lr){11-12}\cmidrule(lr){13-14}\cmidrule(lr){15-16}\cmidrule(lr){17-18}
 & & MSE & MAE & MSE & MAE & MSE & MAE & MSE & MAE & MSE & MAE & MSE & MAE & MSE & MAE & MSE & MAE \\
\midrule
ETTh1 & 96 & 0.382 & 0.401 & \textbf{0.375} & \textbf{0.398} & 0.382 & 0.403 & 0.385 & 0.402 & 0.386 & 0.405 & 0.460 & 0.447 & 0.384 & 0.402 & 0.395 & 0.407 \\
 & 192 & 0.430 & 0.430 & \textbf{0.427} & \textbf{0.429} & 0.429 & 0.435 & 0.443 & 0.430 & 0.441 & 0.436 & 0.512 & 0.477 & 0.436 & \textbf{0.429} & 0.446 & 0.441 \\
 & 336 & \textbf{0.463} & \textbf{0.445} & 0.472 & 0.450 & 0.468 & 0.448 & 0.512 & 0.470 & 0.487 & 0.458 & 0.546 & 0.496 & 0.491 & 0.469 & 0.489 & 0.467 \\
 & 720 & 0.471 & 0.470 & 0.476 & 0.474 & \textbf{0.469} & \textbf{0.461} & 0.497 & 0.476 & 0.503 & 0.491 & 0.544 & 0.517 & 0.521 & 0.500 & 0.513 & 0.510 \\
\addlinespace[1pt]
ETTh2 & 96 & \textbf{0.280} & \textbf{0.331} & 0.293 & 0.345 & 0.286 & 0.338 & 0.289 & 0.342 & 0.297 & 0.349 & 0.308 & 0.355 & 0.340 & 0.374 & 0.340 & 0.394 \\
 & 192 & \textbf{0.357} & \textbf{0.382} & 0.367 & 0.394 & 0.363 & 0.389 & 0.378 & 0.397 & 0.380 & 0.400 & 0.393 & 0.405 & 0.402 & 0.414 & 0.482 & 0.479 \\
 & 336 & \textbf{0.408} & \textbf{0.421} & 0.419 & 0.431 & 0.414 & 0.423 & 0.432 & 0.434 & 0.428 & 0.432 & 0.427 & 0.436 & 0.452 & 0.452 & 0.591 & 0.541 \\
 & 720 & 0.415 & 0.435 & 0.427 & 0.445 & \textbf{0.408} & \textbf{0.432} & 0.464 & 0.464 & 0.427 & 0.445 & 0.436 & 0.450 & 0.462 & 0.468 & 0.839 & 0.661 \\
\addlinespace[1pt]
ETTm1 & 96 & \textbf{0.314} & \textbf{0.348} & 0.326 & 0.364 & 0.318 & 0.356 & 0.317 & 0.356 & 0.334 & 0.368 & 0.352 & 0.374 & 0.338 & 0.375 & 0.346 & 0.374 \\
 & 192 & \textbf{0.361} & \textbf{0.377} & 0.367 & 0.383 & 0.362 & 0.383 & 0.367 & 0.384 & 0.377 & 0.391 & 0.390 & 0.393 & 0.374 & 0.387 & 0.382 & 0.391 \\
 & 336 & 0.393 & 0.408 & 0.402 & 0.409 & 0.395 & 0.407 & \textbf{0.391} & \textbf{0.406} & 0.426 & 0.420 & 0.421 & 0.414 & 0.410 & 0.411 & 0.415 & 0.415 \\
 & 720 & \textbf{0.447} & \textbf{0.440} & 0.469 & 0.446 & 0.452 & 0.441 & 0.454 & 0.441 & 0.491 & 0.459 & 0.462 & 0.449 & 0.478 & 0.450 & 0.473 & 0.451 \\
\addlinespace[1pt]
ETTm2 & 96 & \textbf{0.168} & \textbf{0.254} & 0.178 & 0.260 & 0.171 & 0.256 & 0.175 & 0.257 & 0.180 & 0.264 & 0.183 & 0.270 & 0.187 & 0.267 & 0.193 & 0.293 \\
 & 192 & \textbf{0.230} & \textbf{0.299} & 0.242 & 0.303 & 0.237 & \textbf{0.299} & 0.240 & 0.302 & 0.250 & 0.309 & 0.255 & 0.314 & 0.249 & 0.309 & 0.284 & 0.361 \\
 & 336 & \textbf{0.293} & \textbf{0.336} & 0.303 & 0.342 & 0.296 & 0.338 & 0.303 & 0.343 & 0.311 & 0.348 & 0.309 & 0.347 & 0.321 & 0.351 & 0.382 & 0.429 \\
 & 720 & \textbf{0.384} & 0.399 & 0.400 & 0.399 & 0.392 & \textbf{0.394} & 0.392 & 0.396 & 0.412 & 0.407 & 0.412 & 0.404 & 0.408 & 0.403 & 0.558 & 0.525 \\
\addlinespace[1pt]
Exchange & 96 & \textbf{0.082} & \textbf{0.201} & 0.085 & 0.204 & 0.084 & 0.202 & 0.090 & 0.235 & 0.086 & 0.206 & 0.088 & 0.205 & 0.107 & 0.234 & 0.088 & 0.218 \\
 & 192 & \textbf{0.167} & \textbf{0.292} & 0.178 & 0.299 & 0.187 & 0.307 & 0.187 & 0.343 & 0.177 & 0.299 & 0.176 & 0.299 & 0.226 & 0.344 & 0.176 & 0.315 \\
 & 336 & 0.306 & 0.400 & 0.328 & 0.414 & 0.350 & 0.429 & 0.353 & 0.473 & 0.331 & 0.417 & \textbf{0.301} & \textbf{0.397} & 0.367 & 0.448 & 0.313 & 0.427 \\
 & 720 & \textbf{0.793} & \textbf{0.669} & 0.817 & 0.679 & 1.006 & 0.748 & 0.934 & 0.761 & 0.847 & 0.691 & 0.901 & 0.714 & 0.964 & 0.746 & 0.839 & 0.695 \\
\addlinespace[1pt]
Weather & 96 & \textbf{0.152} & \textbf{0.200} & 0.166 & 0.207 & 0.157 & 0.205 & 0.163 & 0.209 & 0.174 & 0.214 & 0.186 & 0.227 & 0.172 & 0.220 & 0.195 & 0.252 \\
 & 192 & \textbf{0.198} & \textbf{0.244} & 0.216 & 0.254 & 0.204 & 0.247 & 0.211 & 0.254 & 0.221 & 0.254 & 0.234 & 0.265 & 0.219 & 0.261 & 0.237 & 0.295 \\
 & 336 & \textbf{0.251} & \textbf{0.285} & 0.273 & 0.296 & 0.261 & 0.290 & 0.263 & 0.293 & 0.278 & 0.296 & 0.284 & 0.301 & 0.280 & 0.306 & 0.282 & 0.331 \\
 & 720 & \textbf{0.322} & 0.347 & 0.351 & 0.346 & 0.340 & \textbf{0.341} & 0.344 & 0.348 & 0.358 & 0.347 & 0.356 & 0.349 & 0.365 & 0.359 & 0.345 & 0.382 \\
\bottomrule
\end{tabular}}
\end{scriptsize}
\caption{Detailed dataset--horizon comparison against all seven Table~1 baselines. Bold values mark the lowest reported three-decimal error in each row and metric, so MSE and MAE winners may come from different methods. TimeXer ETT/Weather values follow its original report, while its Exchange row follows the aligned input-length-96 reproduction of \cite{naghashi2025should}; other external values follow their recorded aligned sources.}
\label{tab:full_baseline_grid_main}
\end{table*}

\begin{table*}[!t]
\centering
\scriptsize
\setlength{\tabcolsep}{5pt}
\begin{tabular}{llcccc}
\toprule
Dataset & Horizon & AnchorMTS & TimePro & Best MSE & Best MAE \\
\midrule
ETTh1 & 96 & 0.382/0.401 & 0.375/0.398 & TimePro 0.375 & TimePro 0.398 \\
ETTh1 & 192 & 0.430/0.430 & 0.427/0.429 & TimePro 0.427 & TimePro 0.429 \\
ETTh1 & 336 & 0.463/0.445 & 0.472/0.450 & TimeXer 0.468 & TimeXer 0.448 \\
ETTh1 & 720 & 0.471/0.470 & 0.476/0.474 & TimeXer 0.469 & TimeXer 0.461 \\
\midrule
ETTh2 & 96 & 0.280/0.331 & 0.293/0.345 & TimeXer 0.286 & TimeXer 0.338 \\
ETTh2 & 192 & 0.357/0.382 & 0.367/0.394 & TimeXer 0.363 & TimeXer 0.389 \\
ETTh2 & 336 & 0.408/0.421 & 0.419/0.431 & TimeXer 0.414 & TimeXer 0.423 \\
ETTh2 & 720 & 0.415/0.435 & 0.427/0.445 & TimeXer 0.408 & TimeXer 0.432 \\
\midrule
ETTm1 & 96 & 0.314/0.348 & 0.326/0.364 & TimeMixer 0.317 & TimeXer 0.356 \\
ETTm1 & 192 & 0.361/0.377 & 0.367/0.383 & TimeXer 0.362 & TimePro 0.383 \\
ETTm1 & 336 & 0.393/0.408 & 0.402/0.409 & TimeMixer 0.391 & TimeMixer 0.406 \\
ETTm1 & 720 & 0.447/0.440 & 0.469/0.446 & TimeXer 0.452 & TimeXer 0.441 \\
\midrule
ETTm2 & 96 & 0.168/0.254 & 0.178/0.260 & TimeXer 0.171 & TimeXer 0.256 \\
ETTm2 & 192 & 0.230/0.299 & 0.242/0.303 & TimeXer 0.237 & TimeXer 0.299 \\
ETTm2 & 336 & 0.293/0.336 & 0.303/0.342 & TimeXer 0.296 & TimeXer 0.338 \\
ETTm2 & 720 & 0.384/0.399 & 0.400/0.399 & TimeXer 0.392 & TimeXer 0.394 \\
\midrule
Exchange & 96 & 0.082/0.201 & 0.085/0.204 & TimeXer 0.084 & TimeXer 0.202 \\
Exchange & 192 & 0.167/0.292 & 0.178/0.299 & PatchTST 0.176 & TimePro 0.299 \\
Exchange & 336 & 0.306/0.400 & 0.328/0.414 & PatchTST 0.301 & PatchTST 0.397 \\
Exchange & 720 & 0.793/0.669 & 0.817/0.679 & TimePro 0.817 & TimePro 0.679 \\
\midrule
Weather & 96 & 0.152/0.200 & 0.166/0.207 & TimeXer 0.157 & TimeXer 0.205 \\
Weather & 192 & 0.198/0.244 & 0.216/0.254 & TimeXer 0.204 & TimeXer 0.247 \\
Weather & 336 & 0.251/0.285 & 0.273/0.296 & TimeXer 0.261 & TimeXer 0.290 \\
Weather & 720 & 0.322/0.347 & 0.351/0.346 & TimeXer 0.340 & TimeXer 0.341 \\
\bottomrule
\end{tabular}
\caption{Complete fixed-identity AnchorMTS grid and the cellwise best Table~1 references. Best MSE and Best MAE are selected independently from all seven displayed baselines in each cell.}
\label{tab:full_grid_supp}
\end{table*}

\section{Experimental Protocol and Implementation}

\begin{table*}[!t]
\centering
\scriptsize
\begin{tabular}{llrrrrrl}
\toprule
Dataset & Horizon & $d$ & $d_{ff}$ & Layers & Batch & LR & Loss \\
\midrule
ETTh1 & 96 & 512 & 2048 & 2 & 32 & 1.0e-04 & Huber \\
ETTh1 & 192 & 512 & 2048 & 2 & 32 & 1.0e-04 & Huber \\
ETTh1 & 336 & 512 & 2048 & 2 & 32 & 1.0e-04 & Huber \\
ETTh1 & 720 & 128 & 512 & 1 & 16 & 1.0e-04 & MSE \\
ETTh2 & 96 & 512 & 2048 & 2 & 32 & 1.0e-04 & Huber \\
ETTh2 & 192 & 512 & 2048 & 2 & 32 & 1.0e-04 & Huber \\
ETTh2 & 336 & 512 & 2048 & 2 & 32 & 1.0e-04 & Huber \\
ETTh2 & 720 & 512 & 2048 & 2 & 32 & 1.0e-04 & Huber \\
ETTm1 & 96 & 512 & 2048 & 2 & 32 & 1.0e-04 & Huber \\
ETTm1 & 192 & 512 & 2048 & 2 & 32 & 1.0e-04 & Huber \\
ETTm1 & 336 & 128 & 1024 & 1 & 4 & 1.0e-04 & MSE \\
ETTm1 & 720 & 256 & 1024 & 1 & 16 & 2.0e-04 & MSE \\
ETTm2 & 96 & 512 & 2048 & 3 & 32 & 5.0e-05 & Huber \\
ETTm2 & 192 & 512 & 2048 & 2 & 32 & 1.0e-04 & Huber \\
ETTm2 & 336 & 512 & 2048 & 2 & 32 & 1.0e-04 & Huber \\
ETTm2 & 720 & 128 & 512 & 1 & 16 & 1.0e-04 & MSE \\
Exchange & 96 & 256 & 2048 & 1 & 32 & 1.0e-04 & MSE \\
Exchange & 192 & 256 & 1024 & 1 & 16 & 2.0e-04 & MSE \\
Exchange & 336 & 128 & 1024 & 1 & 4 & 1.0e-04 & MSE \\
Exchange & 720 & 256 & 1024 & 1 & 16 & 2.0e-04 & MSE \\
Weather & 96 & 512 & 2048 & 2 & 32 & 1.0e-04 & Huber \\
Weather & 192 & 512 & 2048 & 3 & 32 & 5.0e-05 & Huber \\
Weather & 336 & 512 & 2048 & 2 & 32 & 1.0e-04 & Huber \\
Weather & 720 & 512 & 1024 & 1 & 8 & 5.0e-05 & MSE \\
\bottomrule
\end{tabular}
\caption{Selected ordinary training and temporal-capacity hyperparameters. Every row retains the identical multimodal protocol.}
\label{tab:selected_configs_supp}
\end{table*}

\begin{table*}[!t]
\centering
\scriptsize
\setlength{\tabcolsep}{4pt}
\begin{tabular}{p{1.45cm}p{3.25cm}p{6.1cm}p{3.0cm}}
\toprule
Method & Published regime & Protocol coverage relative to Table~1 & Role here \\
\midrule
VisionTS \cite{chen2024visionts} & Zero-shot plus one-epoch full-shot & Dataset-specific context lengths (576--2304); Exchange absent; no complete aligned $6\times4$ grid. & Visual-forecasting context \\
Time-VLM \cite{siru2025time} & Few-shot, zero-shot, and supervised & Long-term input length 512; Exchange absent; no complete aligned $6\times4$ grid. & Vision--language forecasting context \\
DMMV \cite{shen2026multi} & Full-data supervised & Default look-back 336; Exchange absent; no complete aligned $6\times4$ grid. & Alternative-view fusion context \\
TimesCLIP \cite{dong2025teaching} & Supervised aligned views & Input length 96, but ETTh1/ETTh2 and horizons 336/720 are not reported. & Cross-view alignment context \\
\bottomrule
\end{tabular}
\caption{Published evaluation protocols of the closest multimodal forecasters. Their reported grids differ from Table~1 in context length, dataset coverage, and/or horizon coverage; partial results are therefore not mixed into the 24-setting macro comparison.}
\label{tab:multimodal_protocol_audit}
\end{table*}

\subsection{Dataset and Evaluation Details}

\begin{table}[!t]
\centering
\scriptsize
\setlength{\tabcolsep}{3pt}
\begin{tabular}{lrrl}
\toprule
Dataset & Variables & Frequency & Split \\
\midrule
ETTh1/ETTh2 & 7 & hourly & 12/4/4 months \\
ETTm1/ETTm2 & 7 & 15-minute & 12/4/4 months \\
Exchange & 8 & daily & 70/10/20\% \\
Weather & 21 & 10-minute & 70/10/20\% \\
\bottomrule
\end{tabular}
\caption{Dataset metadata and chronological train/validation/test partitions. Input length is 96 and horizons are 96, 192, 336, and 720.}
\label{tab:dataset_metadata_supp}
\end{table}

Each sample predicts every variable; training-partition statistics standardize validation and test data. External results retain their reported three-decimal precision, which is also used for baseline W/T/L comparisons. Cell-resampled intervals use 100,000 bootstrap samples; matched control intervals resample paired cell units. Exact sign tests are two-sided and exclude ties.

\subsection{Implementation and Selected Configurations}

Each window yields a variable-labelled GAF, a line image, and a recurrence image; all are resized to $224\times224$. Recurrence uses $R=\exp[-d/(\sigma+10^{-6})]$, with positive distances defined by $d>10^{-8}$, $\sigma$ their median, and $\sigma=1$ if no positive distance exists. GPT-4.1-mini receives the historical GAF and the fixed instruction in Section~\ref{sec:caption_protocol}; CLIP ViT-B/32 encodes the images and caption. The shared router applies layer normalization and a scalar linear scorer with bias to each pooled visual representation, then uses temperature 1 and softmax over the three available views before view dropout, without top-$k$ selection. Forecasting uses patch length 16, eight attention heads, and view, image, and text dropout of 0.10, 0.10, and 0.30. The state/output ratios are 0.60, 0.25, and 0.25. The final projection uses $\epsilon=10^{-8}$ with no denominator floor. Optimization and checkpoint scoring use the candidate forecast; final test evaluation returns the projected forecast. Experiments were conducted on an NVIDIA GeForce RTX 4060 Ti using Python 3.8.20, PyTorch 2.4.1 with CUDA 12.1, NumPy 1.24.4, and pandas 2.0.3.

The data loader standardizes each channel using statistics fitted on the training partition. The forecasting model then performs detached per-window mean and variance normalization, decodes anchor and candidate in that normalized state, and applies the same inverse per-window transform to both. A shared learned channel-wise affine transformation, $y\mapsto y[1+0.40\tanh(a_c)]+0.26\tanh(b_c)$, is then applied identically to both forecasts; the exact projection is computed afterward in this common reported space.

View construction, caption generation, and frozen CLIP encoding are preprocessing steps that produce the representations consumed by the forecasting model. We therefore do not interpret cached-representation forecasting latency as end-to-end online inference cost.

\subsection{Hyperparameter Search Scope}

\begingroup
\sloppy
Each dataset--horizon setting evaluates six initial configurations. Eleven settings additionally evaluate a 12-configuration refinement panel using successive halving with epoch budgets 1, 3, and 8 and reduction factor 3, yielding 18 additional training evaluations (12, 4, and 2). The number of evaluated training configurations per setting is therefore $6/6/24$ for the minimum, median, and maximum.
\par
\endgroup

The explored search space spans temporal widths \allowbreak($\{128,\allowbreak256,\allowbreak512\}$), feed-forward widths \allowbreak($\{512,\allowbreak1024,\allowbreak2048\}$), encoder depths \allowbreak($\{1,\allowbreak2,\allowbreak3\}$), batch sizes \allowbreak($\{4,\allowbreak8,\allowbreak16,\allowbreak32\}$), and learning rates \allowbreak($\{5\times10^{-5},\allowbreak10^{-4},\allowbreak2\times10^{-4}\}$). Objectives include MSE and Huber, together with one MSE--MAE configuration with MAE weight 0.1; EMA is either disabled or uses decay 0.99. The panels additionally vary epoch budget and residual-head initialization. The self-derived representation set, frozen encoders, fusion topology, and $(\rho_v,\rho_t,\rho_f)$ remain fixed across all 24 settings.

\subsection{Caption Generation Protocol}
\label{sec:caption_protocol}

\paragraph{Exact fixed instruction.}
\begin{quote}
\scriptsize\raggedright
This is a Gramian Angular Field (GAF) image transformed from a multivariate time series of length 96. The image is a grid of GAF representations for each dimension, with dimension names at the top. In GAF, colors represent the angular sum/difference of time points, where lighter colors (e.g., yellow, red in a 'jet' colormap) indicate larger angular values (similarity), and darker colors (e.g., blue, purple) indicate smaller angular values (dissimilarity). Your task is to generate a concise textual description of the patterns in the image, following this template:\allowbreak{}"A Gramian Angular Field of [N] time series. [Variables with strong periodicity] show strong periodicity, [variables with seasonal or external impact] indicate seasonal impact, [variables with stable trend] remain stable. Overall: [summary of cycles/trend]."\allowbreak{}Requirements: - Keep the output under 77 tokens. - Mention variable names shown in the image. - Highlight differences between variables (periodic, seasonal, stable). - Be compact, clear, and avoid redundant words. Please replace [N] with the actual number of time series dimensions shown in the image, and fill in the bracketed placeholders with specific variable names or descriptions based on your analysis. If a category is not applicable, you can state 'None' or 'No significant [pattern]'. Please provide the description text directly, without any additional prefixes or explanations.\allowbreak{}The following is an example:\allowbreak{}"A Gramian Angular Field of 7 time series. MUFL and LUFL show strong periodicity with distinct block patterns. HUFL, HULL, MULL, LULL indicate seasonal impact with moderate angular variations. OT remains relatively stable with minimal temporal structure. Overall: mixed periodic and seasonal cycles with varying intensity across load variables."
\end{quote}

\paragraph{API fields.}
The implementation calls the chat-completions API with model identifier \texttt{gpt-4.1-mini} and \texttt{max\_tokens=77}. Temperature, \texttt{top\_p}, seed, and response format are not explicitly specified in the request.

\paragraph{Response preprocessing.}
The caption is read from \path{response.choices[0].message.content} and stripped. It is not subsequently truncated as text. During frozen CLIP encoding, the processor uses padding to maximum length, truncation enabled, and \texttt{max\_length=77}, which enforces the CLIP token limit. An API exception writes no CSV row; a later run resumes from image names without valid rows. On resume, prior rows whose stripped caption begins with \texttt{error:} are removed and their images are retried. No retry count or backoff is explicitly specified.

\paragraph{Artifact reproducibility.}
The six dataset caption CSV files are present in the repository and paired with the saved frozen token artifacts. All captions used in the reported experiments are preserved with the frozen artifacts, so reproducing the forecasting results requires neither re-querying the language-model API nor relying on future API-side generation defaults.

\subsection{Closest Multimodal Forecasting Protocols}

Broader multi-view forecasting work spans several fusion objectives. CAMul dynamically weights view-specific information and uncertainty for calibrated probabilistic forecasts \cite{kamarthi2022camul}. TEFN aggregates time- and channel-dimensional evidence through basic probability assignments \cite{zhan2025time}. T3Time aligns time, spectral, and prompt branches with adaptive multi-head fusion \cite{chowdhury2026t3time}, while TriTS combines time, frequency, and visual representations \cite{ao2026trits}. These studies motivate richer multi-view forecasting representations and fusion mechanisms. AnchorMTS focuses on a different design axis: auxiliary representations are allowed to refine a complete numerical forecast, while their final influence is independently constrained in the reported prediction space.

Table~\ref{tab:multimodal_protocol_audit} audits the public paper protocols rather than treating unlike regimes as interchangeable results.

\FloatBarrier
\begin{figure*}[!t]
\section{Representation Analysis}
\subsection{Representation Example}
\centering
\includegraphics[width=0.97\textwidth]{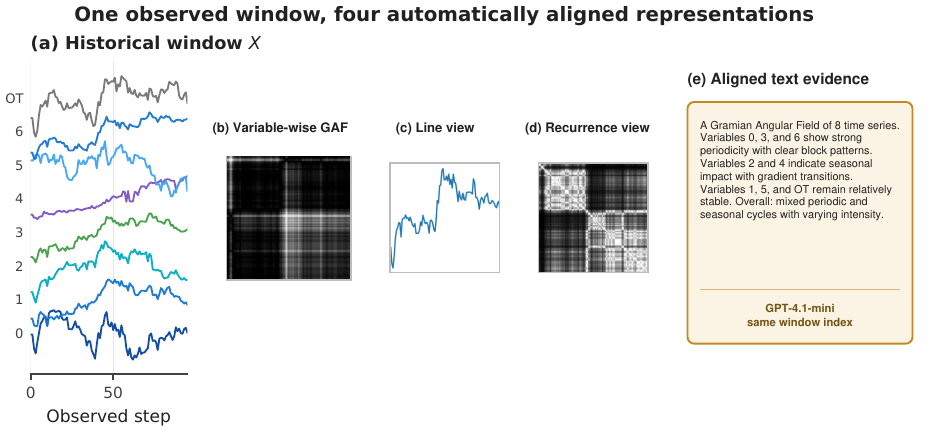}
\caption{Representations generated from one Exchange test history. The variable-wise GAF preserves channel-specific pairwise geometry; line and recurrence views expose trajectory and return structure; the caption summarizes the same observed window. No target-window values are used. The GAF is shown in grayscale for display; caption generation uses the corresponding color rendering.}
\label{fig:evidence-example}
\end{figure*}

\FloatBarrier
\subsection{Retrieval Details}

The main paper reports the macro retrieval lifts; this section gives the descriptor definitions and sampling protocol.

The retrieval analysis uses up to 128 non-overlapping training windows per dataset with stride at least 96. Trend is absolute standardized slope; periodicity is the largest non-DC Fourier amplitude divided by total non-DC amplitude. Recurrence is the dispersion of diagonal means of $\exp(-|s_i-s_j|/(\operatorname{std}|s_i-s_j|+\epsilon))$, and channel diversity is the across-channel dispersion of dominant spectral ratios. Cosine similarity retrieves five neighbors from frozen post-layer-normalized CLIP class tokens. For descriptor $d$, define
\[
\begin{aligned}
D_{\mathcal N}&=\mathbb{E}_{i,j\in\mathcal{N}_5(i)}|d_i-d_j|,\\
D_{\rm all}&=\mathbb{E}_{i<j}|d_i-d_j|,
\end{aligned}
\qquad \mathrm{lift}=1-D_{\mathcal N}/D_{\rm all}.
\]

\section{Reproduction Entry Points}

\path{export_selected_forecast_decompositions.py} and \path{export_final_projected_controls.py} replay checkpoints without training and save candidate, anchor, and final arrays. \path{build_final_result_manifest.py} recomputes projections, metrics, paired intervals, exact sign tests, and violation counts; \path{generate_final_tables.py} and the paper plotting scripts regenerate the reported artifacts. The first-stage, refinement, selected-configuration, caption, and exact-bound manifests retain the source paths and recorded protocol fields used above.

\bibliographystyle{IEEEbib}
\bibliography{references}

\end{document}